\documentclass{article}

\usepackage{times}
\usepackage{helvet} 
\usepackage{courier} 
\usepackage[hyphens]{url}  
\usepackage{graphicx} 
\usepackage{caption}
\DeclareCaptionStyle{ruled}{labelfont=normalfont,labelsep=colon,strut=off} 
\usepackage{algorithm}

\usepackage{amsmath}
\usepackage{amsthm}
\usepackage{amssymb}
\usepackage{bbm}
\usepackage{mathtools}
\usepackage{algpseudocode}
\usepackage{subfigure}

\usepackage[top=30truemm,bottom=30truemm,left=25truemm,right=25truemm]{geometry}

\newtheorem{theorem}{Theorem}
\newtheorem{Lemma}{Lemma}

\newtheorem{proposition}{Proposition}

\DeclareMathOperator*{\argmin}{arg\,min}
\DeclareMathOperator*{\argmax}{arg\,max}
\usepackage{newfloat}
\usepackage{listings}
\floatstyle{ruled}
\newfloat{listing}{tb}{lst}{}
\floatname{listing}{Listing}

\title{\textbf{Partial Wasserstein Covering}}

\author {
\textbf{Keisuke Kawano, Satoshi Koide, Keisuke Otaki} \\
{\normalsize Toyota Central R\&D Labs., Inc.}    \\ 
{\normalsize \{kawano, koide, otaki\}@mosk.tytlabs.co.jp}
}

\date{}

\begin{document}

\maketitle

\begin{abstract}
    We consider a general task called \emph{partial Wasserstein covering} with the goal of providing information on what patterns are not being taken into account in a dataset (e.g., dataset used during development) compared with another dataset(e.g., dataset obtained from actual applications).
    We model this task as a discrete optimization problem with \emph{partial Wasserstein divergence} as an objective function.
    Although this problem is NP-hard, we prove that it satisfies the submodular property, allowing us to use a greedy algorithm with a 0.63 approximation. However, the greedy algorithm is still inefficient because it requires solving linear programming for each objective function evaluation.
    To overcome this inefficiency, we propose \emph{quasi-greedy} algorithms that consist of a series of acceleration techniques,  such as sensitivity analysis based on strong duality and the so-called $C$-transform in the optimal transport field.
    Experimentally, we demonstrate that we can efficiently fill in the gaps between the two datasets and find missing scene in real driving scenes datasets.
\end{abstract}

\section{Introduction}
\label{sec:intro}
A major challenge in real-world machine learning applications is coping with mismatches between the data distribution obtained in real-world applications and those used for development.
% Standard development processes includes choosing models to use, designing subroutines for fail safe, and training/testing models.
Regions in the real-world data distribution that are not well supported in the development data distribution (i.e., regions with low relative densities) result in potential risks such as a lack of evaluation or high generalization error, which in turn leads to low product quality.
Our motivation is to provide developers with information on what patterns are not being taken into account when developing products by selecting some of the (usually \emph{unlabeled}) real-world data distribution, also referred to as \emph{application dataset},\footnote{More precisely, we call a finite set of data sampled from the real-world data distribution \emph{application dataset} in this study.} to fill in the gaps in the \emph{development dataset}.
Note that the term ``development'' includes choosing models to use, designing subroutines for fail safe, and training/testing models.

Our research question is formulated as follows.
To resolve the lack of data density in development datasets, how and using which metric can we select data from application datasets with a limited amount of data for developers to understand?
One reasonable approach is to define the discrepancy between the data distributions and select data to minimize this discrepancy.
The Wasserstein distance has attracted significant attention as a metric for data distributions~\cite{pmlr-v70-arjovsky17a, alvarez2020geometric}.
However, the Wasserstein distance is not capable of representing the \emph{asymmetric} relationship between application datasets and development datasets, i.e., the parts that are over-included during development increase the Wasserstein distance.

\begin{figure}[t]
    \centering
    \includegraphics[width=0.7\linewidth]{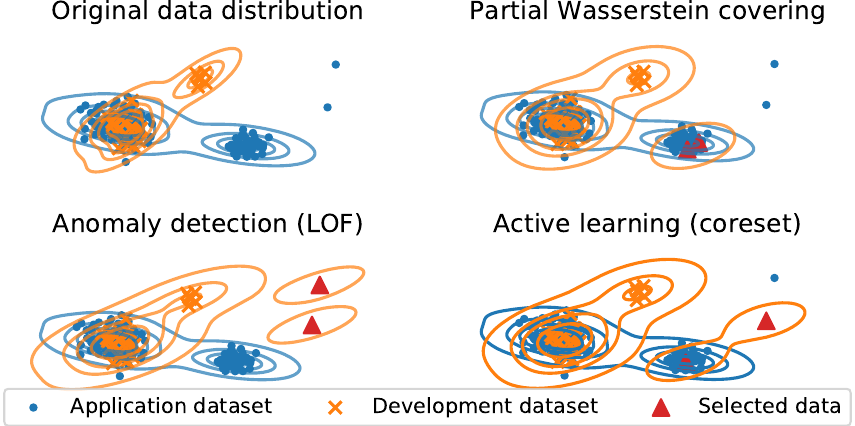}
    \caption{Concept of PWC. PWC extracts some data from an unlabeled application dataset by minimizing the partial Wasserstein divergence between the application dataset and the union of the selected data and a development dataset.
        PWC focuses on regions in which development data are lacking compared with the application dataset, whereas anomaly detection extracts irregular data, and active learning selects to improve accuracy.}
    \label{fig:fig1}
\end{figure}

In this paper, we propose \emph{partial Wasserstein covering (PWC)} that selects a limited amount of data from the application dataset by minimizing the partial Wasserstein divergence~\cite{bonneel2019spot} between the application dataset and the union of the development dataset and the selected data.
PWC, as illustrated in Fig.~\ref{fig:fig1}, selects data from areas with fewer development data than application data in the data distributions (lower-right area in the blue distribution in Fig.~\ref{fig:fig1}) while ignoring areas with sufficient development data (upper-middle of the orange distribution).
We also highlight the data selected through an anomaly detection method, LOF~\cite{breunig2000lof}, where irregular data (upper-right points) were selected, but the major density difference was ignored.
Furthermore, we show the selection obtained using an active learning method, coreset with a $k$-center~\cite{sener2018active}, where
the data are chosen to improve the accuracy rather than fill the gap in terms of the distribution mismatch.

Our main contributions are summarized as follows.
\begin{itemize}
    \item We propose PWC that extracts data that are lacking in a development dataset from an application dataset by minimizing the partial Wasserstein divergence between an application dataset and the development dataset.
    \item We prove that PWC is a maximization problem involving a \emph{submodular} function whose inputs are the set of selected data. This allows us to use a greedy algorithm with a guaranteed approximation ratio.
    \item Additionally, we propose fast heuristics based on sensitivity analysis and the Sinkhorn derivative for an accelerated computation.
    \item Experimentally, we demonstrate that compared with baselines, PWC extracts data lacking in the development data distribution from the application distribution more efficiently.
\end{itemize}

The remainder of this paper is organized as follows.
Section~\ref{sec:preliminary} briefly introduces the partial Wasserstein divergence, linear programming (LP), and submodular functions.
In Sect.~\ref{sec:pwc}, we detail the proposed PWC, fast algorithms for approximating the optimization problem, and some theoretical results.
Section~\ref{sec:related} presents a literature review of related work.
We demonstrated the PWC in some numerical experiments in Sect.~\ref{sec:experiment}.
In Sect.~\ref{sec:conclusion}, we present our conclusions.

\section{Preliminaries}
\label{sec:preliminary}
In this section, we introduce the notations used throughout this paper.
We then describe partial Wasserstein divergences, sensitivity analysis for LP, and submodular functions.
\paragraph{Notations}
Vectors and matrices are denoted in bold (e.g., $\mathbf x$ and $\mathbf A$), where $x_i$ and $A_{ij}$ denote the $i$th and $(i,j)$th elements, respectively. To clarify the elements, we use notations such as $\mathbf x=(f(i))_i$ and $\mathbf A=(g(i,j))_{ij}$, where $f$ and $g$ specify the element values depending on their subscripts.
$\left<\cdot, \cdot\right >$ denotes the inner product of matrices or vectors.
$\mathbbm 1_{n}\in \mathbb R^n$ is a vector with all elements equal to one.
For a natural number $n$, we define $[\![n]\!]\coloneqq\{1,\cdots,n\}$.
For a finite set $V$, we denote its power set as $2^V$ and its cardinality as $|V|$.
The $L_2$-norm is denoted as $\|\cdot\|$.
The delta function is denoted as $\delta(\cdot)$.
$\mathbb R_+$ is a set of real positive numbers, and $[a, b]\subset\mathbb R$ denotes a closed interval.
$I[\cdot]$ denotes an indicator function (zero for false and one for true).

\paragraph{Partial Wasserstein divergence}
\label{sec:pre-pwass}
In this paper, we consider the \emph{partial Wasserstein divergence}~\cite{figalli2010optimal, bonneel2019spot, chapel2020partial} as an objective function.
Partial Wasserstein divergence is designed to measure the discrepancy between two distributions with \emph{different total masses} by considering variations in optimal transport.
Throughout this paper, datasets are modeled as empirical distributions that are represented by mixtures of delta functions without necessarily having the same total mass.
Suppose two empirical distributions (or datasets) $X$ and $Y$ with probability masses $\mathbf a\in\mathbb R^{m}_+$ and $\mathbf b\in\mathbb R^{n}_+$, which are denoted as $X=\sum_{i=1}^{m} a_i\delta(\mathbf x^{(i)})$ and $Y=\sum_{j=1}^{n} b_j\delta(\mathbf y^{(j)})$.
Without loss of generality, the total mass of $Y$ is greater than or equal to that of $X$ (i.e., $\sum_{i=1}^{m} a_i = 1$ and $\sum_{j=1}^{n} b_j \geq 1$).

Based on the definitions above, we define the partial Wasserstein divergence as follows:\footnote{The partial optimal transport problems in the literature contain a wider problem definition than Eq.\eqref{eq:pwass-constraint} as summarized in Table 1~(b) of \cite{bonneel2019spot}, but this paper employs this one-side relaxed Wasserstein divergence corresponding to Table 1~(c) in \cite{bonneel2019spot} without loss of generality. }
\begin{align}
    \label{eq:pwass-constraint}
    \begin{split}
        &\mathcal {PW}^2(X, Y) \coloneqq  \!\!\min_{\mathbf P\in U(\mathbf a, \mathbf b)}\!\! \left< \mathbf P,  \mathbf C\right>,\text{where}  \\   &
        \ U(\mathbf a, \mathbf b) = \{ \mathbf P \in [0,1]^{m \times n} \ | \ \mathbf P \mathbbm 1_{n}= \mathbf a, \ \mathbf P\!^\top\! \mathbbm 1_{m} \leq \mathbf b \},
    \end{split}
\end{align}
where $C_{ij}\coloneqq \|\mathbf x^{(i)}-\mathbf y^{(j)}\|^2$ is the transport cost between $\mathbf x^{(i)}$ and $\mathbf y^{(j)}$, and $P_{ij}$ is the amount of mass flowing from $\mathbf x^{(i)}$ to $\mathbf y^{(j)}$ (to be optimized).
Unlike the standard Wasserstein distance, the second constraint in Eq.~\eqref{eq:pwass-constraint} is not defined with ``$=$'', but with ``$\leq$''.
This modification allows us to treat distributions with different total masses.
The condition $\mathbf P^\top \mathbbm 1_{m} \leq \mathbf b$ indicates that the mass in $Y$ does not need to be transported, whereas the condition $\mathbf P \mathbbm 1_{n}= \mathbf a$ indicates that all of the mass in $X$ should be transported without excess or deficiency (just as in the original Wasserstein distance).
This property is useful for the problem defined below, which treats datasets with vastly different sizes.

To compute the partial Wasserstein divergence, we must solve the minimization problem in Eq.\eqref{eq:pwass-constraint}.
In this paper, we consider the following two methods.
(i) LP using simplex method.
(ii) Generalized Sinkhorn iteration with entropy regularization (with a small regularization parameter $\varepsilon>0$)~\cite{benamou2015iterative, peyre2019computational}.

As will be detailed later, an element in mass $\mathbf b$ varies when adding data to the development datasets.
A key to our algorithm is to quickly estimate the extent to which the partial Wasserstein divergence will change when an element in $\mathbf b$ varies.
If we compute $\mathcal{PW}^2$ using LP, we can employ a \emph{sensitivity analysis},  which will be described in the following paragraph.
If we use generalized Sinkhorn iterations, we can use automatic differentiation techniques to obtain a partial derivative with respect to~$b_j$ (see Sect.~\ref{sec:algorithms}).

\paragraph{LP and sensitivity analysis}
\label{sec:prelim:LP}
The sensitivity analysis of LP plays an important role in our algorithm.
Given a variable $\mathbf{x} \in \mathbb R^m$ and parameters $\mathbf c\in \mathbb{R}^m$, $\mathbf{d}\in \mathbb R^n$, and $\mathbf A \in \mathbb R^{n \times m}$, the standard form of LP can be written as follows: $\min\ \mathbf c^\top\mathbf{x},\  \text{s.t.} \  \mathbf A\mathbf{x} \leq \mathbf{d}, \ \mathbf{x} \geq 0.$
Sensitivity analysis is a framework for estimating changes in a solution when the parameters $\mathbf c, \mathbf A$, and $\mathbf d$ of the problem vary.
We consider a sensitivity analysis for the right-hand side of the constraint (i.e., $\mathbf d$).
When $d_j$ changes as $d_j+\Delta d_j$, the optimal value changes by $\mathbf{y}^*_j \Delta  d_j$ if a change $\Delta d_j$ in $d_j$ lies within $(\underline{d}_j,\overline{d}_j)$, where $\mathbf y^*$ is the optimal solution of the following dual problem corresponding to the primal problem:
$\max \ \mathbf d^\top\mathbf{y} \ \text{s.t. }\ \mathbf A^\top\mathbf{y} \geq \mathbf{c},\ \mathbf{y} \geq 0.$
We refer readers to \cite{vanderbei2015linear} for the details of calculating the upper bound $\overline{d}_j$ and the lower bound $\underline {d}_i$.

\paragraph{Submodular function}
Our covering problem is modeled as a discrete optimization problem involving \emph{submodular functions}, which are a subclass of set functions that play an important role in discrete optimization.
A set function $\phi: 2^V\to \mathbb R$ is called a \emph{submodular} iff $\phi(S\cup T) + \phi(S\cap T) \leq \phi(S) + \phi(T) \ (\forall S, T \subseteq V).$
A submodular function is \emph{monotone} iff $\phi(S)\leq \phi(T)$ for $S\subseteq T$.
An important property of monotone submodular functions is that a greedy algorithm provides a $(1 -1/e) \approx 0.63$-approximate solution to the maximization problem under a budget constraint $|S|\leq K$~\cite{nemhauser1978analysis}.
The \emph{contraction} $\tilde{\phi} : 2^V\to\mathbb{R}$ of a (monotone) submodular function $\phi$, which is defined as $\tilde{\phi}(S) \coloneqq  \phi(S\cup T) - \phi(T)$, where $T\subseteq V$, is also a (monotone) submodular function.

\section{Partial Wasserstein covering problem}
\label{sec:pwc}

\subsection{Formulation}
\label{subsec:formulation}
As discussed in Sect.~\ref{sec:intro}, our goal is to fill in the gap between the application dataset $\mathcal D_\text{app}$ by adding some data from a candidate dataset $\mathcal D_\text{cand}$ to a small dataset $\mathcal D_\text{dev}$.
We consider $\mathcal D_\text{cand} = \mathcal D_\text{app}$ (i.e., we copy some data in $\mathcal D_\text{app}$ and add them into $D_\text{dev}$) in the above-mentioned scenarios, but we herein consider the most general formulation.
We model this task as a discrete optimization problem called the \emph{partial Wasserstein covering problem}.

Given a dataset for development $\mathcal D_\text{dev} = \{\mathbf y^{(j)}\}_{j=1}^{N_\text{dev}}$, a dataset obtained from an application $\mathcal D_\text{app} = \{\mathbf x^{(i)}\}_{i=1}^{N_\text{app}}$, and a dataset containing candidates for selection $\mathcal D_\text{cand} = \{\mathbf s^{(j)}\}_{j=1}^{N_\text{cand}}$, where  $N_\text{app} \geq N_\text{dev}$, the PWC problem is defined as the following optimization:\footnote{We herein consider the partial Wasserstein divergence between the unlabled datasets, because the application and candidate datasets are usually unlabeled. If they are labeled, we can use the labels information as in \cite{alvarez2020geometric}.}

\begin{equation}
    \label{eq:selection}
    \max_{\substack{S \subseteq \mathcal D_\text{cand}\\\text{s.t. }|S|\leq K}}- \mathcal{PW}^2(\mathcal D_\text{app}, S \cup \mathcal D_\text{dev}) +  \mathcal{PW}^2(\mathcal D_\text{app}, \mathcal D_\text{dev})
\end{equation}

We select a subset $S$ from the candidate dataset $\mathcal D_\text{cand}$ under the budget constraint $|S|\leq K~(\ll N_\text{cand})$, and then add that subset to the development data $\mathcal D_\text{dev}$ to minimize the partial Wasserstein divergence between the two datasets $\mathcal D_\text{app}$ and $S \cup \mathcal D_\text{dev}$.
The second term is a constant with respect to $S$, which is included to make the objective non-negative.

In Eq.~\eqref{eq:selection}, we must specify the probability mass (i.e., $\mathbf a$ and $\mathbf b$ in Eq.~\eqref{eq:pwass-constraint}) for each data point.
Here, we employ a uniform mass, which is a natural choice because we do not have prior information regarding each data point.
Specifically, we set the weights to $a_i=1/N_\text{app}$ for $\mathcal D_\text{app}$ and $b_j=1/N_\text{dev}$ for $S\cup\mathcal D_\text{dev}$.
With this choice of masses, we can easily show that $\mathcal{PW}^2(\mathcal D_\text{app}, S\cup\mathcal D_\text{dev}) = \mathcal{W}^2(\mathcal D_\text{app}, \mathcal D_\text{dev})$ when $S=\emptyset$, where $\mathcal{W}^2(\mathcal D_\text{app}, \mathcal D_\text{dev})$ is the Wasserstein distance. Therefore, based on the monotone property (Sect.~\ref{sec:method:proof}), the objective value is non-negative for any $S$.

The PWC problem in Eq.\eqref{eq:selection} can be written as a mixed integer linear program (MILP) as follows.
Instead of using the divergence between $\mathcal D_\text{app}$ and $S\cup\mathcal D_\text{dev}$, we consider the divergence between $\mathcal D_\text{app}$ and $\mathcal D_\text{cand}\cup\mathcal D_\text{dev}$.
Hence, the mass $\mathbf b$ is an $(N_\text{cand}+N_\text{dev})$-dimensional vector, where the first $N_\text{cand}$ elements correspond to the data in $\mathcal D_\text{cand}$ and the remaining elements correspond to the data in $\mathcal D_\text{dev}$.
In this case, we never transport to data points in $\mathcal D_\text{cand}\setminus S$, meaning we use the following mass that depends on $S$:
\begin{equation}
    \label{eq:parameterized_b}
    b_j(S)  = \begin{cases}
        \frac{I[\mathbf s^{(j)} \in S]}{N_\text{dev}} , & \text{if  } 1\leq j \leq N_\text{cand} \\
        \frac{1}{N_\text{dev}},                         & \text{if  } N_\text{cand} + 1\leq j    % \leq N_\text{cand} + N_\text{dev}
        .
    \end{cases}
\end{equation}
As a result, the problem is an MILP problem with an objective function $\langle\mathbf P, \mathbf C\rangle$ w.r.t. $S\subseteq\mathcal D_\text{cand}$ and $\mathbf P\in U(\mathbf a, \mathbf b(S))$ with $|S|\leq K$ and $a_i=1/N_\text{app}$.

One may wonder why we use the partial Wasserstein divergence in Eq.\eqref{eq:selection} instead of the standard Wasserstein distance.
This is because the asymmetry in the conditions of the partial Wasserstein divergence enables us to extract only the parts with a density lower than that of the application dataset, whereas the standard Wasserstein distance becomes large for parts with sufficient density in the development dataset.
Furthermore, the guaranteed approximation algorithms that we describe later can be utilized only for the partial Wasserstein divergence version.

\subsection{Submodularity of the PWC problem}
\label{sec:method:proof}
In this section, we prove the following theorem to guarantee the approximation ratio of the proposed algorithms.
\begin{theorem}
    \label{eq:theorem}
    Given the datasets $X=\{\mathbf x^{(i)}\}_{i=1}^{N_x}$ and $Y=\{\mathbf y^{(j)}\}_{j=1}^{N_y}$, and a subset of a dataset $S \subseteq \{\mathbf s^{(j)}\}_{j=1}^{N_s}$,
    $\phi(S)= -\mathcal{PW}^2(X, S\cup Y) + \mathcal {PW}^2(X, Y)$ is a monotone submodular function.
\end{theorem}

To prove Theorem~\ref{eq:theorem}, we reduce our problem to the partial maximum weight matching problem~\cite{bar2016tight}, which is known to be submodular.
First, we present the following lemmas.

\begin{Lemma}
    \label{lemma:rational}
    %Let $\mathbb Q$ be a set of rational numbers, and suppose $A \in \mathbb Q^{n\times m}$ and $\mathbf b\in \mathbb Q^{m}$.
    Let $\mathbb Q$ be the set of all rational numbers, and $m$ and $n$ be positive integers with $n\leq m$.
    Consider $\mathbf A \in \mathbb Q^{m\times n}$ and $\mathbf b\in \mathbb Q^{m}$.
    Then, the extreme points $\mathbf x^*$ of a convex polytope defined by the linear inequalities $\mathbf A\mathbf x \leq \mathbf b$ are also rational, meaning $\mathbf x^*\in \mathbb Q^{n}$.
\end{Lemma}
\textbf{Proof of Lemma~\ref{lemma:rational}.}
It is well known~\cite{vanderbei2015linear} that any extreme point of the polytope is obtained by (i) choosing $n$ inequalities from $\mathbf A\mathbf x\le \mathbf b$ and (ii) solving the corresponding $n$-variable linear system $\mathbf B\mathbf x=\tilde{\mathbf b}$, where $\mathbf B$ and $\tilde{\mathbf b}$ are the corresponding sub-matrix/vector (when replacing $\leq$ with $=$).
As all elements in $\mathbf B^{-1}$ and $\tilde{\mathbf b}$ are rational,
\footnote{This immediately follows from the fact that $\mathbf B^{-1}=\tilde {\mathbf B}/\det {\mathbf B}$, where $\tilde {\mathbf B}$ is the adjugate matrix of $\mathbf B$ and $\det\mathbf{B}$ is the determinant of $\mathbf{B}$. Since $\mathbf B\in\mathbb Q^{n\times n}$, $\tilde {\mathbf B}$ and $\det \mathbf B$ are both rational.}
we conclude that any extreme points are rational. \qed

\begin{Lemma}
    \label{lemma:prop_appendix}
    Let $l, m$, and $n$ be positive integers and $Z=[\![n]\!]$. Given a positive-valued $m$-by-$n$ matrix $\mathbf R>0$, the following set function $\psi:2^Z\to\mathbb R$ is a submodular function:

    \begin{align}
        \begin{split}
            \psi(S) =& \max_{\mathbf P\in U^{\leq}(\mathbbm{1}_m/m, \mathbf{b}(S))} \left<\mathbf R, \mathbf P\right>, \\
            &\text{where}\ \  b_j(S) = \tfrac{I[j\in S]}{l} \quad \forall j\in[\![n]\!].
        \end{split}
    \end{align}
    Here, $U^{\leq}(\mathbf{a}, \mathbf{b}(S))$ is a set defined by replacing the constraint $\mathbf P \mathbbm{1}_n = \mathbf{a}$ in Eq.~(1) with $\mathbf P \mathbbm{1}_n \leq \mathbf{a}$.
\end{Lemma}
\textbf{Proof of Lemma~\ref{lemma:prop_appendix}.}
For any fixed $S\subseteq Z$, there is an optimal solution $\mathbf P^{*}$ of $\psi(S)$, where all elements $P_{ij}^{*}$ are rational because all the coefficients in the constraints are in $\mathbb Q$ ($\because$~Lemma~\ref{lemma:rational}).
Thus, there is a natural number $M \in \mathbb Z_+$ that satisfies
$\psi(S) = \frac{1}{mlM} \max_{\mathbf P\in U_{\mathbb{Q}}(S)} \left<\mathbf R,\mathbf P\right>$
where the set $U_{\mathbb{Q}}(S)$ is the set of possible transports when the transport mass is restricted to the rational numbers $\mathbb{Q}$.
\begin{align}
    \begin{split}
        U_{\mathbb{Q}}(S) = \left\{\mathbf P\in \mathbb Z_+^{m \times n}  \mid  \mathbf P\mathbbm 1_n \leq lM\cdot\mathbbm 1_m, \right.  \\
        \left. \mathbf P^\top\mathbbm 1_m \leq mM\cdot I[j\in S]\cdot\mathbbm{1}_n\right\}.
    \end{split}
\end{align}

The problem of $\psi(S)$ can be written as a network flow problem on a bipartite network following Sect~3.4 in \cite{peyre2019computational}.
Since the elements of $\mathbf{P}$ are integers, we can reformulate the problem of $\psi(S)$
using maximum-weight bipartite matching (i.e., an assignment problem) on a bipartite graph whose vertices are generated by copying the vertices $lM$ and $mM$ times with modified edge weights.
Therefore, $G(V\cup W, E)$ is a bipartite graph, where $V=\{v_{1,1},\dots, v_{1,lM}, \dots, v_{m, 1}, \dots, v_{m, lM} \}$ and $W=\{w_{1,1},\dots, w_{1,mM}, \dots, w_{n, 1}, \dots, w_{n, mM}\}$.
For any $i\in[\![m]\!], j\in[\![n]\!]$, the transport mass $P_{ij}\in\mathbb{Z}_+$ is encoded by using the copied vertices $\{v_{i,1},\dots,v_{i,lM}\}$ and $\{w_{j,1},\dots, w_{j,mM}\}$, and the weights of the edges among them are $\frac{1}{mlM}R_{ij} > 0$ to represent $R_{ij}$.
The partial maximum weight matching function for a graph with positive edge weights is a submodular function~\cite{bar2016tight}.
Therefore, $\psi(S)$ is a submodular function. \qed

\begin{Lemma}
    \label{lemma:constraints_compatibility}
    If $|S| \geq l$, there exists an optimal solution $\mathbf P^{*}$ in the maximization problem of $\psi$ satisfying $\mathbf P^{*}\mathbbm{1}_n = \frac{\mathbbm 1_m}{m}$ (i.e., $\mathbf{P}^* \coloneqq  \argmax_{\mathbf{P}\in U^{\leq} ( \frac{\mathbbm 1_m}{m}, \mathbf b(S))} \left<\mathbf R, \mathbf P\right> = \argmax_{\mathbf{P}\in U(\frac{\mathbbm 1_m}{m}, \mathbf b(S))} \left<\mathbf R,\mathbf P\right>$).
\end{Lemma}
\textbf{Proof of Lemma~\ref{lemma:constraints_compatibility}.}
Given an optimal solution $\mathbf P^*$, suppose there exists $i'$ satisfying $\sum_{j=1}^n P^*_{i'j} < \frac{1}{m}$.
This implies that $\sum_{i,j} P^*_{ij}<1$.
We denote the gap as $\delta\coloneqq \frac{1}{m} - \sum_{j=1}^{n}  P_{i'j}^*>0$.
Based on the assumption of $|S|\geq l$, we have $\sum_{j=1}^{n}  b_j(S) \geq 1$.
Hence, there must exist $j'$ such that $\sum_{i=1}^{m}  P_{ij'}^* <  b_{j'}$.
We denote the gap as $\delta'\coloneqq  b_{j'} - \sum_{i=1}^{m}  P_{ij'}^* > 0$.
We also define $\tilde \delta = \min \{\delta, \delta'\} > 0$.
Then, a matrix $\mathbf P'=(P_{ij}^*+\tilde\delta I[i=i', j=j'])_{ij}$ is still a feasible solution.
However, $R_{ij}>0$ leads to $\left<\mathbf R, \mathbf P'\right > > \left< \mathbf R, \mathbf P^*\right >$, which contradicts $\mathbf P^*$ being the optimal solution.
Thus, $\sum_{j=1}^{n} P_{ij}^* = \frac{1}{m}, \forall i$. \qed

\textbf{Proof of Theorem~\ref{eq:theorem}.}
Given $C_\text{max} = \max_{i,j}  C_{ij} + \gamma$, where $\gamma>0$, the following is satisfied:
\begin{align}
    \begin{split}
        \phi(S)
        & = (-C_\text{max} + \max_{\mathbf P\in U(\mathbf a, \mathbf b(S))} \left < \mathbf R, \mathbf P\right>) \\
        &\ \  - (-C_\text{max} + \max_{\mathbf P\in U(\mathbf a, \mathbf b(\emptyset))} \left< \mathbf R, \mathbf P\right >),
    \end{split}
\end{align}
where $\mathbf R = (C_\text{max} -  C_{ij})_{ij} > 0$, $m=N_x$, $n = N_s + N_y$, and $l=N_y$.
Here, $|S\cup Y| \geq N_y = l$ and Lemma \ref{lemma:constraints_compatibility} yield $\phi(S) = \psi(S \cup Y) - \psi(Y)$.
Since $\phi(S)$ is a contraction of the submodular function $\psi(S)$ (Lemma~\ref{lemma:prop_appendix}), $\phi(S) = -\mathcal{PW}^2(X, S\cup Y) + \mathcal {PW}^2(X,Y)$ is a submodular function.

We now prove the monotonicity of $\phi$.
For $S\subseteq T$, we have $U(\mathbf{a},\mathbf{b}(S))\subseteq U(\mathbf{a}, \mathbf{b}(T))$ because $\mathbf b(S)\leq \mathbf b(T)$.
This implies that $\phi(S)\leq \phi(T)$.
\qed

Finally, we prove Proposition~\ref{prop:1} to clarify the computational intractability of the PWC problem.
\begin{proposition}
    \label{prop:1}
    The PWC problem with marginals $\mathbf{a}$ and $\mathbf{b}(S)$ is NP-hard.
\end{proposition}
\textbf{Proof of Proposition~\ref{prop:1}.}
The decision version of the PWC problem, denoted as $\mathtt{PWC}(\mathbf{a},\mathbf{b}(S),C,K,x)$, asks whether a subset $S, |S|\leq K$ such that $\phi(S)\leq x\in\mathbb{R}$ exists.
We discuss a reduction to the \texttt{PWC} from the well-known NP-complete problems called set cover problems~\cite{karp1972reducibility}.
Given a family $\mathcal{G}=\{G_j\subseteq \mathcal{I}\}_{j=1}^{M}$ from the set $\mathcal{I}=[\![N]\!]$ and an integer $k$, $\mathtt{SetCover}(\mathcal{G},\mathcal{I}, k)$ is a decision problem for answering $\mathtt{Yes}$ iff there exists a sub-family $\mathcal{G'}\subseteq\mathcal{G}$ such that $|\mathcal{G'}|\leq k$ and $\cup_{G\in\mathcal{G'}}G=\mathcal{I}$.
Our reduction is defined as follows.
Given $\mathtt{SetCover}(\mathcal{G}, \mathcal{I}, k)$ for $i\in [\![N]\!]$ and $j\in [\![M]\!]$, we set $a_i=1, b_j(S)=|G_j|\cdot I[j\in S],  C_{ij}=I[i\in G_j], K=k$, and $x=0$.
Based on this polynomial reduction, the given $\mathtt{SetCover}$ instance is $\texttt{Yes}$ iff the reduced $\mathtt{PWC}$ instance is $\mathtt{Yes}$.
This reduction means that the decision version of \texttt{PWC} is NP-complete and
now Proposition~\ref{prop:1} holds.

\subsection{Algorithms}
\label{sec:algorithms}
\paragraph{MILP and greedy algorithms}
The PWC problem in Eq.~\eqref{eq:selection} can be solved directly as an MILP problem.
We refer to this method as \texttt{PW-MILP}.
However, this is extremely inefficient when $N_\text{app}$, $N_\text{dev}$, and $N_\text{cand}$ are large.

As a consequence of Theorem~\ref{eq:theorem}, a simple greedy algorithm provides a 0.63 approximation because the problem is a type of monotone submodular maximization with a budget constraint $|S|\leq K$~\cite{nemhauser1978analysis}.
%For the monotone submodular maximization problem with budget constraint $|S|\leq K$, a simple greedy algorithm achieves 0.63-approximation~\cite{nemhauser1978analysis}.
Specifically, this greedy algorithm selects data from the candidate dataset $\mathcal D_\text{cand}$ sequentially in each iteration to maximize $\phi$.
% minimize the partial Wasserstein divergence (i.e., $\argmax_{\mathbf{s}^{(j)}\notin S} \phi(\{\mathbf{s}^{(j)}\} \cup S) = \argmin_{\mathbf{s}^{(j)} \notin S}\phi(\{\mathbf{s}^{(j)}\}\cup S)$.
We consider two baseline methods, called \texttt{PW-greedy-LP} and \texttt{PW-greedy-ent}.
\texttt{PW-greedy-LP} solves the LP problem using the simplex method. \texttt{PW-greedy-ent} computes the optimal $\mathbf{P}^\star$ using Sinkhorn iterations~\cite{benamou2015iterative}.

Even for the greedy algorithms mentioned above, the computational cost is still high because we need to calculate the partial Wasserstein divergences for all candidates on $\mathcal D_\text{app}$ in each iteration, yielding a computational time complexity of $O(K\!\cdot\!N_\text{cand} \!\cdot\!C_{PW})$.
Here, $C_{PW}$ is the complexity of the partial Wasserstein divergence with the simplex method for the LP or Sinkhorn iteration.

To reduce the computational cost, we propose heuristic approaches called \emph{quasi-greedy algorithms}.
A high-level description of the quasi-greedy algorithms is provided below.
In each step of greedy selection, we use a type of \emph{sensitivity value} that allows us to \emph{estimate} how much the partial Wasserstein divergence varies when adding candidate data, instead of performing the actual computation of divergence.
Specifically, if we add a data point $\mathbf{s}^{(j)}$ to $S$, denoted as $T=\{\mathbf s^{(j)}\} \cup S$, we have $\mathbf b(T)=\mathbf b(S)+\frac{\mathbf e_j}{N_\text{dev}}$, where $\mathbf e_j$ is a one-hot vector with the corresponding $j$th element activated.
Hence, we can estimate the change in the divergence for data addition by computing the sensitivity with respect to $\mathbf b(S)$.
If efficient sensitivity computation is possible, we can speed up the greedy algorithms.
We describe the concrete methods for each of the approaches, simplex method, and Sinkhorn iterations in the following paragraphs.

\paragraph{Quasi-greedy algorithm for the simplex method}
When we use the simplex method for the partial Wasserstein divergence computation, we employ a sensitivity analysis for LP.
The dual problem of partial Wasserstein divergence computation is defined as follows.
\begin{align}
    \label{eq:dualwass}
    \begin{split}
        \max_{\mathbf f \in \mathbb R^{N_\text{app}}, \mathbf g \in \mathbb R^{N_\text{cand}+N_\text{dev}}} \left <\mathbf f, \mathbf a \right > + \left <\mathbf g,\mathbf b(S) \right >, \\
        \text{s.t.\ \ }  g_j \leq 0, \ f_i +  g_j \leq  C_{ij}, \forall i,j,
    \end{split}
\end{align}
where $\mathbf f$ and $\mathbf g$ are dual variables.
Using sensitivity analysis, the changes in the partial Wasserstein divergences can be estimated as $g_j^* \Delta  b_j$, where $g_j^*$ is an optimal dual solution of Eq.~\eqref{eq:dualwass}, and $\Delta  b_j$ is the change in $b_j$.
It is worth noting that $\Delta b_j$ now corresponds to $I[\mathbf{s}^{(j)}\in S]$ in Eq.~\eqref{eq:parameterized_b}, and a smaller Eq.~\eqref{eq:dualwass} results in a larger Eq.~\eqref{eq:selection}.
Thus, we propose a heuristic algorithm that iteratively selects $\mathbf s^{(j^*)}$ satisfying $j^* = \argmin_{j\in[\![N_\text{cand}]\!], s^{(j)}\notin S} \ g_j^{*(t)}$ at iteration $t$, where $g_j^{*(t)}$ is the optimal dual solution at~$t$.
We emphasize that we have $- \frac{1}{N_\text{dev}} g_j^*=\phi(\{\mathbf s^{(j)}\} \cup S) - \phi(S) $ as long as $\overline{b}_j - b_j \geq 1/N_\text{dev}$ holds, where $\overline{b}_j$ is the upper bound obtained from the sensitivity analysis, leading to the same selection as that of the greedy algorithm.
The computational complexity of the heuristic algorithm is $O(K\!\cdot\!C_{PW})$ because we can obtain $\mathbf g^*$ by solving the dual simplex.
It should be noted that we add a small value to $b_j$ when $b_j=0$ to avoid undefined values in $\mathbf{g}$.
We refer to this algorithm as \texttt{PW-sensitivity-LP} and summarize the overall algorithm in Algorithm~\ref{alg:greedy_sensitivity}.
\begin{algorithm}[H]
    \caption{Data selection for the PWC problem with sensitivity analysis (\texttt{PW-sensitivity-LP})}\label{alg:greedy_sensitivity}
    \begin{algorithmic}[1]
        \State{\textbf{Input: } $\mathcal D_\text{app}, \mathcal D_\text{dev}, \mathcal D_\text{cand}$}
        \State{\textbf{Output: } $S$}
        \State $S \gets \{\}$, $t\gets 0$
        \While{$|S|<K$}
        \State Calculate $\mathcal {PW}^2(\mathcal D_\text{app}, S \cup \mathcal D_\text{dev})$ %(or $\mathcal {PW}^2_\epsilon(\mathcal D_\text{app}, S \cup \mathcal D_\text{dev})$)
        , and obtain $g_j^{*(t)}$ from the sensitivity analysis. % (or derivative $\partial \mathcal {PW}^2_\epsilon(\mathcal D_\text{app}, S \cup \mathcal D_\text{dev})/\partial b_j$).
        \State $j^* = \argmin_{j\in[\![N_\text{cand}]\!], \mathbf s^{(j)}\notin S} \ g_j^{*(t)}$
        %\\( or  $j^* = \argmin_{j\in[\![N_\text{cand}]\!], s^{(j)}\notin S} \ \frac{\partial \mathcal {PW}^2_\epsilon(\mathcal D_\text{app}, S \cup \mathcal D_\text{dev})}{\partial b_j}$)
        \State $S \gets S \cup \{\mathbf{s}^{(j^*)}\}$
        \State $t \gets t+1$
        \EndWhile
    \end{algorithmic}
\end{algorithm}

For computational efficiency, we use the solution matrix $\mathbf P$ from the previous step for the initial value in the simplex algorithm in the current step.
Any feasible solution $\mathbf P \in U(\mathbf{a}, \mathbf{b}(S))$ is also feasible because $\mathbf P \in U(\mathbf{a}, \mathbf{b}(T))$ for all $S \subseteq T$.
The previous solutions of the dual forms $\mathbf f$ and $\mathbf g$ are also utilized for the initialization of the dual simplex and Sinkhorn iterations.

\subparagraph{Faster heuristic using C-transform}
The heuristics above still require solving a large LP in each step, where the number of dual constraints is $N_\text{app}\times (N_\text{cand}+N_\text{dev})$.
Here, we aim to reduce the size of the constraints in the LP to $N_\text{app}\times (|S|+N_\text{dev})\ll N_\text{app}\times (N_\text{cand}+N_\text{dev})$ using \emph{C-transform} (Sect.3.1 in \cite{peyre2019computational}).

To derive the algorithm, we consider the dual Eq.\eqref{eq:dualwass}.
Considering the fact that $b_j(S)=0$ for $\mathbf s^{(j)}\notin S$, we first solve the smaller LP problem by ignoring $j$ such that $\mathbf s^{(j)}\notin S$, whose system size is $N_\text{app}\times (|S|+N_\text{dev})$.
Let the optimal dual solution of this smaller LP be $(\mathbf f^*, \mathbf g^*)$, where $\mathbf f^* \in \mathbb R^{N_\text{app}}, \mathbf g^* \in \mathbb R^{|S| + N_\text{dev}}$.
For each $\mathbf s^{(j)}\notin S$, we consider an LP in which $\mathbf s^{(j)}$ is added to $S$. %, (i.e., $b_j(S)$ changes from zero to $1/N_\text{dev}$ in the original LP).
Instead of solving each LP such as \texttt{PW-greedy-LP}, we approximate the optimal solution using a technique called the $C$-transform.
More specifically, for each $j$ such that $\mathbf s^{(j)}\notin S$, we only optimize $g_j$ and fix the other variables to be the dual optimal above.
This is done by $g_j^{\mathbf C}\coloneqq \min\{0, \min_{i\in[\![N_\text{dev}]\!]} C_{ij}- f_i^*\}$.
Note that this is the largest value of $g_j$, satisfying the dual constraints.
This $g_j^{\mathbf C}$ gives the estimated increase in the objective function when $\mathbf s^{(j)}\notin S$ is added to $S$.
Finally, we select the instance $j^*=\arg\min_{j\in[\![N_\text{cand}]\!], \mathbf s^{(j)}\notin S} g^{\mathbf C}_j$.
As shown later, this heuristic experimentally works efficiently in terms of computational times.
We refer to this algorithm as \texttt{PW-sensitivity-Ctrans}.

\paragraph{Quasi-greedy algorithm for Sinkhorn iteration}
When we use the generalized Sinkhorn iterations, we can easily obtain the derivative of the entropy-regularized partial Wasserstein divergence $\nabla_j\coloneqq \partial \mathcal{PW}^2_\epsilon(\mathcal D_\text{app}, S \cup \mathcal D_\text{dev})/\partial b_j$, where $\mathcal{PW}^2_\epsilon$ denotes the partial Wasserstein divergence, using automatic differentiation techniques such as those provided by PyTorch~\cite{pytorch}.
Based on this derivative, we can derive a quasi-greedy algorithm for the Sinkhorn iteration as $j^* = \argmin_{j\in[\![N_\text{cand}]\!],\mathbf{s}^{(j)}\notin S} \nabla_j$ (i.e., we replace Lines~6 and~7 of Algorithm~\ref{alg:greedy_sensitivity} with this formula).
Because the derivative can be obtained with the same computational complexity as the Sinkhorn iterations, the overall complexity is $O(K\cdot C_{PW})$.
We refer this algorithm \texttt{PW-sensitivity-ent}.

\section{Related work}
\label{sec:related}
\paragraph{Instance selection and data summarization}
Tasks for selecting an important portion of a large dataset are common in machine learning.
\emph{Instance selection} involves extracting a subset of a training dataset while maintaining the target task performance.
Accoring to~\cite{olvera2010review}, two types of instance selection methods have been studied: model-dependent methods~\cite{hart1968condensed,ritter1975algorithm, chou2006generalized,wilson1972asymptotic, vazquez2005stochastic} and label-dependent methods~\cite{wilson2000reduction, brighton2002advances, riquelme2003finding, bezdek2001nearest, liu2002issues, spillmann2006transforming}.
Unlike instance selection methods, PWCs do not depend on either model or ground-truth labels.
\emph{Data summarization} involves finding representative elements in a large dataset~\cite{ahmed2019data}.
%With the advance of ML methodologies, data summarization modeled as optimization problems has been attracted much attention, and the submodularity plays an important role for this approach for various datasets~\cite{mitrovic2018data,lin2011class,mirzasoleiman2016fast,tschiatschek2014learning}.
Here, the submodularity plays an important role~\cite{mitrovic2018data,lin2011class,mirzasoleiman2016fast,tschiatschek2014learning}.
Unlike data summarization, PWC focuses on filling in gaps between datasets.

\paragraph{Anomaly detection}
Our covering problem can be considered as the task of selecting a subset of a dataset that is not included in the development dataset.
In this regard, our method is related to anomaly detectors e.g., LOF~\cite{breunig2000lof}, one-class SVM~\cite{scholkopf1999support}, and deep learning-based methods~\cite{kim2019rapp, an2015variational}.
However, our goal is to extract patterns that are less included in the development dataset, but have a certain volume in the application, rather than selecting rare data that would be considered as observation noise or infrequent patterns that are not as important as frequent patterns.

\paragraph{Active learning}
The problem of selecting data from an unlabeled dataset is also considered in active learning.
A typical approach usees the output of the target model e.g., the entropy-based method~\cite{holub2008entropy, coletta2019combining}, whereas some methods are independent of the output of the model e.g., the $k$-center-based~\cite{sener2018active}, and Wasserstein distance-based approach~\cite{shui2020deep}.
In contrast, our goal is finding unconsidered patterns during development, even if they do not contribute to improving the prediction performance by directly adding them to the training data; developers may use these unconsidered data to test the generalization performance, design subroutines to process the patterns, redevelop a new model, or alert users to not use the products in certain special situations.
Furthermore, we emphasize that the submodularity and guaranteed approximation algorithms can be used only with the partial Wasserstein divergence, but not with the vanilla Wasserstein distance.

\paragraph{Facility location problem (FLP) and $k$-median}
The FLP and related \emph{$k$-median}~\cite{revelle1970central} are similar to our problem in the sense that we select data $S\subseteq\mathcal D_\text{cand}$ to minimize an objective function.
FLP is a mathematical model for finding a desirable location for facilities (e.g., stations, schools, and hospitals).
While FLP models facility opening costs, our data covering problem does not consider data selection costs, making it more similar to $k$-median problems.
However, unlike the $k$-median, our covering problem allows relaxed transportation for distribution, which is known as  \emph{Kantorovich relaxation} in the optimal transport field~\cite{peyre2019computational}.
Hence, our approach using partial Wasserstein divergence enables us to model more flexible assignments among distributions beyond na\"ive assignments among data.

\section{Experiments}
In our experiments, we considered a scenario in which we wished to select some data in the application dataset $\mathcal D_\text{app}$ for the PWC problem (i.e., $\mathcal D_\text{app} = \mathcal D_\text{cand}$).
%We implemented the solvers for the PWC problem as follows.
For \texttt{PW-*-LP} and \texttt{PW-sensitivity-Ctrans}, we used IBM ILOG CPLEX 20.01~\cite{studio2013cplex} for the dual simplex algorithm, where the sensitivity analysis for LP is also available.
For \texttt{PW-*-ent}, we computed the entropic regularized version of the partial Wasserstein divergence using PyTorch~\cite{pytorch} on a GPU.
We computed the sensitivity using automatic differentiation.
We set the coefficient for entropy regularization $\epsilon = 0.01$ and terminated the Sinkhorn iterations when the divergence did not change by at least $\max_{i,j} C_{ij} \times 10^{-12}$ compared with the previous step, or the number of iterations reached 50,000.
All experiments were conducted with an Intel\textregistered ~Xeon\textregistered~Gold 6142 CPU and an NVIDIA\textregistered~TITAN RTX\texttrademark~GPU.

\subsection{Comparison of algorithms}
\label{sec:heuristiceval}
\begin{figure*}[t]
    \centering
    \includegraphics[width=0.98\linewidth]{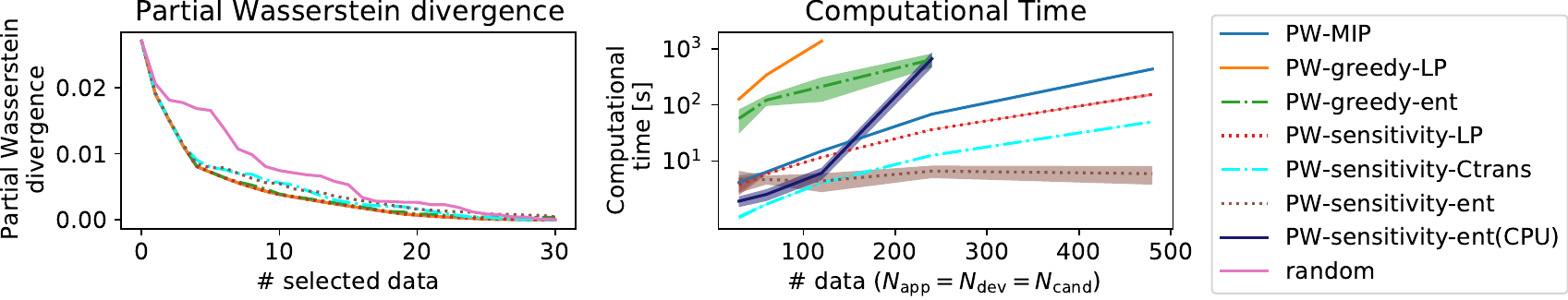}
    \caption{Partial Wasserstein divergence $\mathcal {PW}^2(\mathcal D_\text{app}, S \cup \mathcal D_\text{dev})$  when data are sequentially selected and added to $S$ (left). Computational time, where colored areas indicate standard deviations (right).}
    \label{fig:twofigures}
\end{figure*}

\begin{figure}[t]
    %\begin{minipage}{0.48\hsize}
    \centering
    \includegraphics[width=0.45\linewidth]{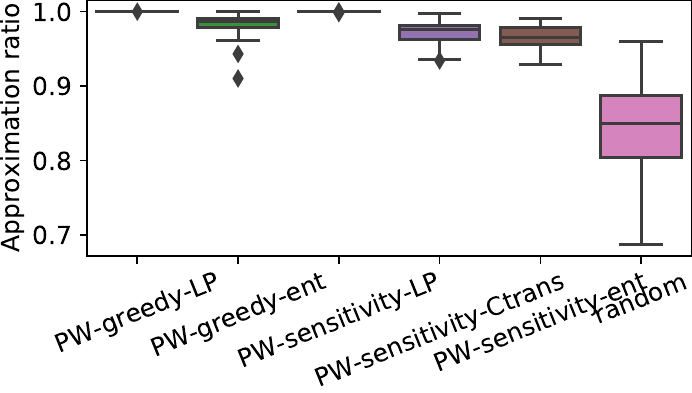}
    \caption{Approximation ratios evaluated empirically for $K=15$ and $50$ random instances.}
    \label{fig:optimal-ratios}
    %\end{minipage}
\end{figure}

First, we compare our algorithms, namely, the greedy-based \texttt{PW-greedy-*}, sensitivity-based \texttt{PW-sensitivity-*}, and random selection for the baseline.
For the evaluation, we generated two-dimensional random values following a normal distribution for $\mathcal D_\text{app}$ and $\mathcal D_\text{dev}$, respectively.
Figure~\ref{fig:twofigures}~(left) presents the partial Wasserstein divergences between $\mathcal D_\text{app}$ and $S \cup \mathcal D_\text{dev}$ while varying the number of the selected data when $N_\text{app} = N_\text{dev} = 30$ and Fig.~\ref{fig:twofigures}~(right) presents the computational times when $K=30$ and $N_\text{app} = N_\text{dev} = N_\text{cand}$ varied.
As shown in Fig.~\ref{fig:twofigures}~(left), \texttt{PW-greedy-LP}, and \texttt{PW-sensitivity-LP} select exactly the same data as the global optimal solution obtained by \texttt{PW-MILP}.
As shown in Fig.~\ref{fig:twofigures}~(right), considering the logarithmic scale, \texttt{PW-sensitivity-*} significantly reduces the computational time compared with \texttt{PW-MILP}, while the na\"ive greedy algorithms do not scale.
In particular, \texttt{PW-sensitivity-ent} can be calculated quickly as long as the GPU memory is sufficient, whereas its CPU version (\texttt{PW-sensitivity-ent(CPU)}) is significantly slower.
\texttt{PW-sensitivity-Ctrans} is the fastest among the methods without GPUs.
It is 8.67 and 3.07 times faster than \texttt{PW-MILP} and \texttt{PW-sensitivity-LP}, respectively.
For quantitative evaluation, we define the \emph{empirical approximation ratio} as $\phi(S_K)/\phi(S^*_K)$, where $S^*_K$ is the optimal solution obtained by \texttt{PW-MILP} when $|S| = K$.
Figure~\ref{fig:optimal-ratios} shows the empirical approximation ratio ($N_\text{app} = N_\text{dev} = 30, K=15$) for 50 different random seeds.
Both \texttt{PW-*-LP} achieve approximation ratios close to one, whereas \texttt{PW-*-ent} and \texttt{PW-sensitivity-Ctrans} have slightly lower approximation ratios\footnote{An important feature of \texttt{PW-*-ent} is that they can be executed without using any commercial solvers.}.

\label{sec:experiment}

\subsection{Finding a missing category}
\label{sec:mnist}
For the quantitative evaluation of the missing pattern findings, we herein consider a scenario in which a category (i.e., label) is less included in the development dataset than in the application dataset\footnote{Note that in real scenarios, missing patterns do not always correspond to labels.}.
We employ subsets of the MNIST dataset~\cite{lecun-mnisthandwrittendigit-2010}, where the development dataset contains $0$ labels at a rate of 0.5\%, whereas all labels are included in the application data in equal ratios (i.e., 10\%).
We randomly sampled 500 images from the validation split for each $\mathcal D_\text{app}$ and $\mathcal D_\text{dev}$.
We employ the squared $L^2$ distance on the pixel space as the ground metric for our method.
For baseline methods, we employed LOF~\cite{breunig2000lof} novelty detection provided by scikit-learn~\cite{scikit-learn} with default parameters.
For the LOF, we selected top-$K$ data according to the anomaly scores.
We also employed active learning approaches based on entropies of the prediction~\cite{holub2008entropy, coletta2019combining} and coreset with the $k$-center and $k$-center greedy~\cite{sener2018active, bachem2017practical}.
For the entropy-based method, we train a LeNet-like neural network using the training split, and then select $K$ data with high entropies of the predictions from the application dataset.
LOF and coreset algorithms are conducted on the pixel space.

\begin{figure}[t]
    %\begin{minipage}{0.48\hsize}
    \centering
    \includegraphics[width=.7\linewidth]{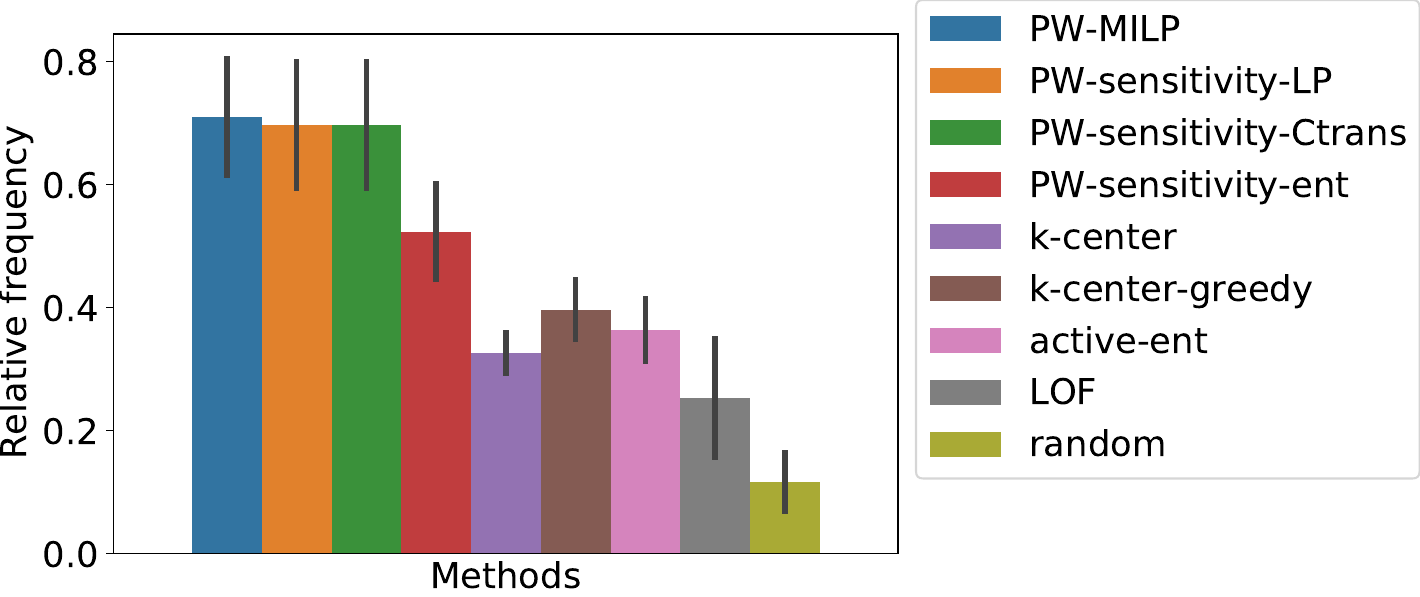}
    \caption{Relative frequency of missing category in selected data. The black bars denote the standard deviations.}
    \label{fig:missing}
    %\end{minipage}
\end{figure}

We conducted 10 trials using different random seeds for the proposed algorithms and baselines, except for the \texttt{PW-greedy-*} algorithms, because they took over 24h.
Figure~\ref{fig:missing} presents a histogram of the selected labels when $K=30$, where $x$-axis corresponds to labels of selected data (i.e., $0$ or not $0$) and $y$-axis shows relative frequencies of selected data.
The proposed methods (i.e., \texttt{PW-*}) extract more data corresponding to the label $0$ (0.71 for \texttt{PW-MILP}) than the baselines.
We can conclude that the proposed methods successfully selected data from the candidate dataset $\mathcal D_\text{cand} (=\mathcal D_\text{app})$ to fill in the missing areas in the distribution.

\subsection{Missing scene extraction in driving datasets}
\label{sec:driving}
\begin{figure}[t]
    \centering
    \subfigure[Partial Wassersein covering]{
        \includegraphics[width=0.7\linewidth]{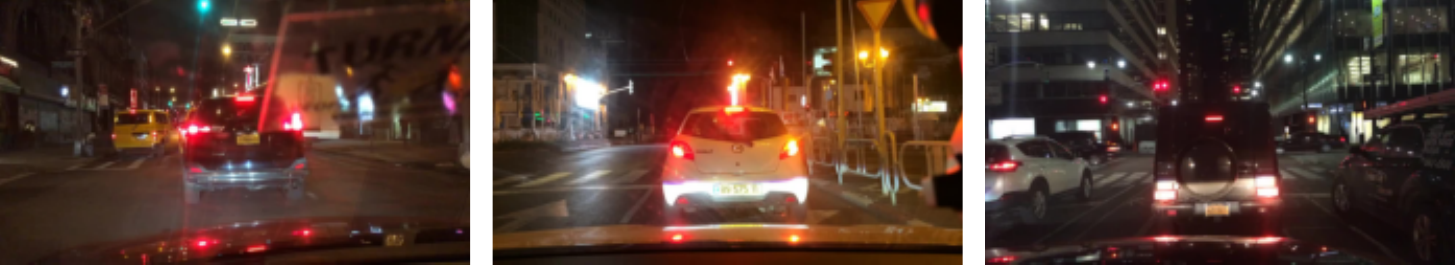}
    }
    \\
    \subfigure[LOF]{
        \includegraphics[width=0.7\linewidth]{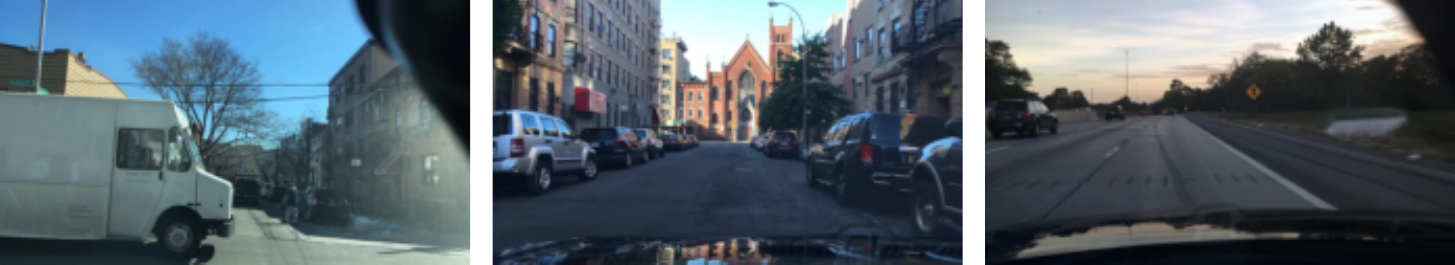}
    } \\
    \subfigure[Coreset ($k$-center greedy)]{
        \includegraphics[width=0.7\linewidth]{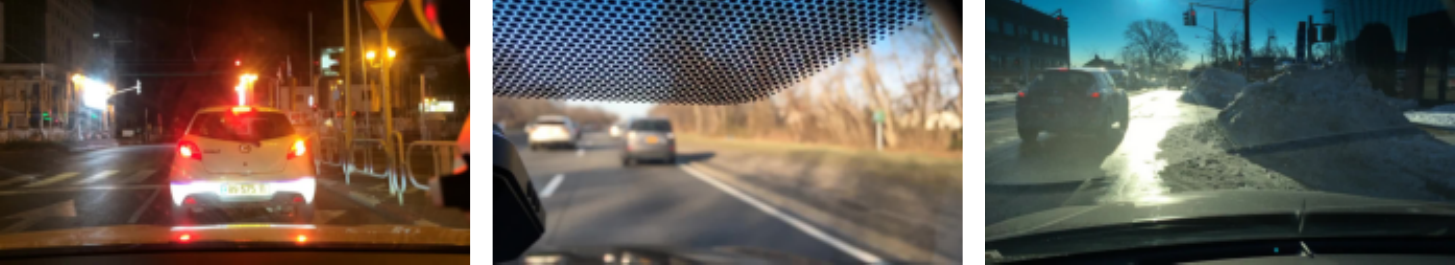}
    } %\\
    % \subfigure[Random]{
    % \includegraphics[width=0.85\linewidth]{./random-3.pdf}
    %     }
    \caption{Covering results when the application dataset (BDD100k) contains nighttime scenes, but the development dataset (KITTI) does not. PWC (\texttt{PW-sensitivity-LP}) extracts the major difference between the two datasets (i.e., nighttime images), whereas LOF and coreset ($k$-center greedy) mainly extract some rare cases.}
    \label{fig:bdd}
\end{figure}

Finally, we demonstrate PWC in a realistic scenario, using driving scene images.
We adopted two datasets, BDD100K~\cite{bdd100k} and KITTI~(Object Detection Evaluation 2012)~\cite{Geiger2013IJRR} as the application and development datasets, respectively.
The major difference between these two datasets is that KITTI ($\mathcal D_\text{dev}$) contains only daytime images, whereas BDD100k ($\mathcal D_\text{app}$) contains both daytime and nighttime images.
To reduce computational time, we randomly selected 1,500 data points for the development dataset from the test split of the KITTI dataset and 3,000 data points for the application dataset from the test split of BDD100k.
To compute the transport costs between images, we calculated the squared $L_2$ norms between the feature vectors extracted by a pretrained ResNet with 50 layers~\cite{he2016deep} obtained from Torchvision~\cite{marcel2010torchvision}.
Before inputting the images into the ResNet, each image was resized to a height of 224 pixels and then center-cropped to a width of 224, followed by normalization.
As baseline methods, we adopted the LOF~\cite{breunig2000lof} and coreset with the $k$-center greedy~\cite{sener2018active}.
Figure~\ref{fig:bdd} presents the obtained top-3 images.
One can see that PWC (\texttt{PW-sensitivity-LP}) selects the major pattern (i.e., nighttime scenes) that is not included in the development data, whereas LOF and coreset mainly extracts specific rare scenes (e.g., a truck crossing a street or out-of-focus scene).
The coreset selects a single image of nighttime scenes; however, we emphasize that providing multiple images is essential for the developers to understand what patterns in the image (e.g., nighttime, type of cars, or roadside) are less included in the image.

The above results indicate that the PWC problem enables us to accelerate updating ML-based systems when the distributions of application and development data are different because our method does not focus on isolated anomalies but major missing patterns in the application data, as shown in Fig.~\ref{fig:fig1} and Fig.~\ref{fig:bdd}.
The ML workflow using our method can efficiently find such patterns and allow developers to incrementally evaluate and update the ML-based systems (e.g., test the generalization performance, redevelop a new model, and designing a special subroutine for the pattern).
Identifying patterns that are not taken into account during development and addressing them individually can improve the reliability and generalization ability of ML-based systems, such as image recognition in autonomous vehicles.

\section{Conclusion}
\label{sec:conclusion}
In this paper, we proposed the PWC, which fills in the gaps between development datasets and application datasets based on partial Wasserstein divergence.
We also proved the submodularity of the PWC, leading to a greedy algorithm with the guaranteed approximation ratio.
In addition, we proposed quasi-greedy algorithms based on sensitivity analysis and derivatives of Sinkhorn iterations.
Experimentally, we demonstrated that the proposed method covers the major areas of development datasets with densities lower than those of the corresponding areas in application datasets.
The main limitation of the PWC is scalability;
the space complexity of the dual simplex or Sinkhorn iteration is at least $O(N_\text{app} \cdot N_\text{dev})$, which might require approximation, e.g., stochastic optimizations~\cite{aude2016stochastic}, slicing~\cite{bonneel2019spot} and neural networks~\cite{xie2019scalable}).
As a potential risk, if infrequent patterns in application datasets are as important as frequent patterns, our proposed method may ignore some important cases, and it may be desirable to use our covering with methods focusing on fairness.

{\small
\bibliography{arxiv}
\bibliographystyle{unsrt}
}

\appendix

\clearpage
\section{Reformulation of Eq.(2) as a MILP}
The MILP formulation of Eq.(2) with marginal masses $\mathbf{a}$ and $\mathbf{b}(S)$ can be defined based on the partial Wasserstein divergence in Eq.(1), i.e., the marginal mass $\mathbf{b}(S)$ depends on which subset $S\subseteq\mathcal{D}_\mathrm{cand}$ is selected. Precisely, the marginal mass of $\mathcal{D}_\mathrm{app}$ side (i.e., source) is $a_i=\frac{1}{N_\mathrm{app}}$. The mass of $\mathcal{D}_\mathrm{cand}\cup\mathcal{D}_\mathrm{dev}$ side (i.e., target) is now Eq.(3), meaning that $b_j(S)$ represents the mass can be transported into $j$. This parameterization enables us to define Eq.(2) as MILP.

Let $y_j\in\{0,1\}, j\in [\![N_\mathrm{cand}]\!]$ be the $0$-$1$ decision variable. The value $y_j=1$ means the data $s^{(j)}\in \mathcal{D}_\mathrm{cand}$ is included in $S$ (i.e., $s^{(j)}\in S$), and $y_j=0$ means $s^{(j)}\notin S$. Continuous decision variables $P_{ij}\in [0,1]$ represent transportation masses between $\mathcal{D}_\mathrm{app}$ and $\mathcal{D}_\mathrm{cand}\cup\mathcal{D}_\mathrm{dev}$ for each data pair $i\in [\![N_\mathrm{app}]\!]$ and $j\in [\![N_\mathrm{cand}+N_\mathrm{dev}]\!]$. The minimization problem among two sets with marginal masses $\mathbf{a}$ and $\mathbf{b}(S)$ is written as the following MILP problem:

\begin{align}
    \min_{\mathbf{P}\in [0,1]^{N_\mathrm{app}\times (N_\mathrm{cand} + N_\mathrm{dev})}, \{y_j\}_{j=1}^{N_\mathrm{cand}}} \langle \mathbf{P}, \mathbf{C}\rangle
\end{align}
subject to

\begin{align}
    \label{eq:st2}
    \sum_{j \in [\![N_\mathrm{cand} + N_\mathrm{dev}]\!]} P_{ij} & = \frac{1}{N_\mathrm{app}}      & \text{ for all }i\in [\![N_\mathrm{app}]\!]                                              \\
    \label{eq:st3}
    \sum_{i \in [\![N_\mathrm{app}]\!]}
    P_{ij}                                                       & \leq \frac{y_j}{N_\mathrm{dev}} & \text{ for all }1\leq j\leq N_\mathrm{app}                                               \\
    \label{eq:st4}
    \sum_{i \in [\![N_\mathrm{app}]\!]}
    P_{ij}                                                       & \leq \frac{1}{N_\mathrm{dev}}   & \text{ for all }N_\mathrm{app}+1\leq j\leq N_\mathrm{app}+N_\mathrm{dev}                 \\
    \label{eq:st5}
    \sum_{j \in [\![N_\mathrm{cand}]\!]} y_j                     & \leq K                                                                                                                     \\
    \label{eq:st6}
    y_j                                                          & \in \{0,1\}                     & \text{ for all }j\in [\![N_\mathrm{cand}]\!]                                             \\
    \label{eq:st7}
    P_{ij}                                                       & \in[0,1]                        & \text{ for all }i\in [\![N_\mathrm{app}]\!], j\in [\![N_\mathrm{cand}+N_\mathrm{dev}]\!]
\end{align}

Constraints \eqref{eq:st2}-\eqref{eq:st4} ensures the requirements of the partial Wasserstein divergence of Eq.~(1). Note that Constraints \eqref{eq:st3} and \eqref{eq:st5} represent that we select up to $K$ data from $\mathcal{D}_\mathrm{cand}$. Constraints \eqref{eq:st6}-\eqref{eq:st7} define decision variables.

\section{Pseudocodes of proposed algorithms}

This section explains the evaluated methods in the main text; the greedy-based algorithm \texttt{PW-greedy-*}, sensitivity-based ones \texttt{PW-sensitivity-*}, and two baselines \texttt{random}.

\subsection{Greedy-based algorithms}

We explain the proposed greedy-based algorithm \texttt{PW-greedy-LP}, which is based on LP formulation of the partial Wasserstein divergence.
The pseudocode is given below.
For each insertion iteration up to $K$, we compute all the partial Wasserstein divergences when $\mathbf{s}^{(j)}$ for $j\in [\![N_\text{cand}]\!]$ only if $\mathbf{s}^{(j)}\notin S$.
We select a new data $\mathbf{s}^{(j)}$ if this addition maximizes the selection objective (i.e., minimize the distance among two datasets).

\begin{algorithm}[H]
    \caption{Greedy data selection for the PWC problem (\texttt{PW-greedy-LP})}
    \begin{algorithmic}[1]
        \State{\textbf{Input: } $\mathcal D_\text{app}, \mathcal D_\text{dev}, \mathcal D_\text{cand}$}
        \State{\textbf{Output: } $S$}
        \State $S \gets \{\}$
        \While{$|S|<K$}
        \Comment {Iteration up to $K$ data
        \State Compute $d_j \coloneqq \mathcal{PW}^2(\mathcal D_\text{app}, S\cup\{\mathbf{s}^{(j)}\}\cup\mathcal D_\text{dev})$ for each $j\in [\![N_\text{cand}]\!]$ only if $\mathbf{s}^{(j)}\notin S$

        \Comment {$O(N_\text{cand}\cdot C_\mathit{PW})$ in each iteration}}

        \Comment {(1). each divergence requires $O(C_\mathit{PW})$}

        \Comment {(2). $N_\text{cand}$ times to compute $\{d_j\}$}

        \State Select $j^\star = \argmin_{j\in [\![N_\text{cand}]\!], \mathbf{s}^{(j)}\notin S} d_j$
        \State $S \gets S \cup \{\mathbf{s}^{(j)}\}$
        \EndWhile
    \end{algorithmic}
    \label{alg:greedy}
\end{algorithm}
Note that \texttt{PW-greedy-ent} is obtained by just replacing the LP part of computing $\mathcal{PW}^2(\cdot, \cdot)$ with the (partial) Sinkhorn iterations of computing the entropy regularized Partial Wasserstein divergence $\mathcal{PW}^2_\epsilon(\cdot, \cdot)$.

\subsection{$C$-transform-based sensitivity algorithm}

Developed sensitivity-based algorithms (e.g., \texttt{PW-sensitivity-LP}) reduces computational costs to evaluate all the divergence $d_{j}$ in greedy-based algorithms to estimate the importance of $j$-th data by their sensitivity values.
In the LP-based method, such values are obtained by sensitivity analyses functions included in the LP solver.
Further, gradients of the partial Wasserstein divergences can be applicable when the entropy-regularized PWC is used.
Note that these methods are explained in the main text (Algorithm~\ref{alg:greedy}).

Here, we give further details particularly when the $C$-transform is adopted to accelerate sensitivity-based algorithms.
The pseudocode is given below.
For each iteration up to $K$, we compute the dual optimal solutions $\mathbf{f}^\star, \mathbf{g}^\star$ with LP-formulation of system size $N_\text{cand} \times (|S|+N_\text{dev})$, instead of using the full size $N_\text{cand} \times (N_\text{dev}+N_\text{dev})$ for practical acceleration.
Once dual optimal solutions are computed by a solver, we could estimate the effect of adding a new data $\mathbf{s}^{(j)}$ using the computed solutions as they are also feasible even if $\mathbf{s}^{(j)}$ is added to the next step LP formulation.
The effect of adding $\mathbf{s}^{(j)}$ can be estimated by $C$-transform as we discussed (see Chapter 3 in \cite{peyre2019computational} as well) and this estimation is apparently possible in constant time (see Line~6).
In conclusion, this $C$-transform-based algorithm has the same computational complexity with those of \texttt{PW-sensitivity-LP}, this computational trick accelerates the total computation times as we evaluated in experiments.

\begin{algorithm}[H]
    \caption{Data selection for the PWC problem with sensitivity analysis and $C$-transform (\texttt{PW-sensitivity-Ctrans})}\label{alg:ctrans}
    \begin{algorithmic}[1]
        \State{\textbf{Input: } $\mathcal D_\text{app}, \mathcal D_\text{dev}, \mathcal D_\text{cand}$}
        \State{\textbf{Output: } $S$}
        \State $S \gets \{\}$
        \While{$|S| < K$}
        \State Compute $\mathcal {PW}^2(\mathcal D_\text{app}, S \cup \mathcal D_\text{dev})$ with LP of system size $N_\text{cand} \times (|S|+N_\text{dev})$, and obtain the optimal dual $\mathbf{f}^\star, \mathbf{g}^\star$, where $\mathbf{f}^\star \in \mathbb{R}^{N_\text{app}}, \mathbf{g}^\star \in \mathbb{R}^{|S| + N_\text{dev}}$
        \State Compute $C$-transforms: For each $j\in[\![N_\text{cand}]\!]$ if $\mathbf{s}^{(j)}$, we evaluate the effect of adding $\mathbf{s}^{(j)}$ with the value $g_j^\mathbf{C}\coloneqq \min\{0, \min_{i\in [\![N_\text{dev}]\!]} C_{i,j} - f_i^*\}$.
        \State $j^\star = \argmin_{j\in [\![N_\text{cand}]\!], \mathbf{s}^{(j)}\notin S} \mathbf{g}^\mathbf{C}_j$
        \State $S \gets S \cup \{\mathbf{s}^{(j^\star)}\}$
        \EndWhile
    \end{algorithmic}
\end{algorithm}

\subsection{Sinkhorn algorithms}
The entropy regularized Wasserstein divergence \cite{cuturi2013sinkhorn} has attracted much attention because it can be computed in high speed on GPUs by some simple matrix multiplier, named the Sinkhorn iterations.
Entropy regularized Wasserstein distance is formally defined as
\begin{align}
    \mathcal{W}^2_\epsilon(X, Y) & =  \min_{\mathbf P\in U(\mathbf a, \mathbf b)} \left<\mathbf P,\mathbf C\right> - \epsilon H(\mathbf{P}),                            \\
    U(\mathbf a, \mathbf b)      & = \{ \mathbf P \in [0,1]^{m \times n} \ | \ \mathbf P \mathbbm 1_{n}= \mathbf a, \ \mathbf P\!^\top\! \mathbbm 1_{m} = \mathbf b \}, \\
    H(\mathbf{P})                & = -\sum_{i=1}^{m} \sum_{j=1}^{n} P_{ij} (\log P_{ij}-1),
\end{align}
where $\epsilon > 0$.
An important property of the Sinkhorn iterations is differentiable with respect to the inputs and constraints, for example, $\partial \mathcal W^2_\epsilon (X,Y) / \partial X$ and $\partial \mathcal W^2_\epsilon (X,Y) / \partial b_j$ can be obtained by the automatic differentiation techniques.
It is noteworthy that we add a small value to $b_j$ for the computation of the derivative.
We can also employ Sinkhorn-like iterations on GPUs to solve the entropy regularized partial Wasserstein divergences~\cite{benamou2015iterative}.

The (partial) derivatives $\partial \mathcal W^2_\epsilon (X,Y) / \partial X$ and $\partial \mathcal W^2_\epsilon (X,Y) / \partial b_j$ can be computed after evaluating $\mathcal{PW}^2_\epsilon$.
For example in \emph{PW-greedy-LP}, we can replace $\mathcal{PW}^2$ with $\mathcal{PW}^2_\epsilon$, leading another proposed method \emph{PW-greedy-ent}.
Further, instead of selecting $j^\star=\arg\min_{j\in [\![N_\mathrm{cand}]\!],\mathbf{s}^{(j)}\notin{}S} d_j$, where $d_j := \mathcal{PW}^2(\mathcal{D}_\mathrm{app}, S\cup\{\mathbf{s}^{(j)}\}\cup\mathcal{D}_\mathrm{dev})$, we could select $j^\star$ that minimizes the partial derivatives of $\mathcal{PW}^2$, leading another proposed method \emph{PW-sensitivity-ent}.

\section{Details of Experiments}

\subsection{LeNet-like network}
We show the PyTorch code of LeNet-like network used for the entropy-based active learning approach in MNIST experiment.
The network is trained by SGD optimizer, where learning rate is 0.001 and momentum is 0.9 to minimize  the cross entropy loss function during 20 epochs.

\begin{lstlisting}
class Net(nn.Module):
    def __init__(self):
        super(Net, self).__init__()
        self.conv1 = nn.Conv2d(1, 32, 3) # 28x28x32 -> 26x26x32
        self.conv2 = nn.Conv2d(32, 64, 3) # 26x26x64 -> 24x24x64 
        self.pool = nn.MaxPool2d(2, 2) # 24x24x64 -> 12x12x64
        self.dropout1 = nn.Dropout2d()
        self.fc1 = nn.Linear(12 * 12 * 64, 128)
        self.dropout2 = nn.Dropout2d()
        self.fc2 = nn.Linear(128, 10)

    def forward(self, x):
        x = F.relu(self.conv1(x))
        x = self.pool(F.relu(self.conv2(x)))
        x = self.dropout1(x)
        x = x.view(-1, 12 * 12 * 64)
        x = F.relu(self.fc1(x))
        x = self.dropout2(x)
        x = self.fc2(x)
        return x
\end{lstlisting}

\subsection{Finding a completely missing category}

\begin{figure}[t]
    \centering
    \includegraphics[width=0.46\linewidth]{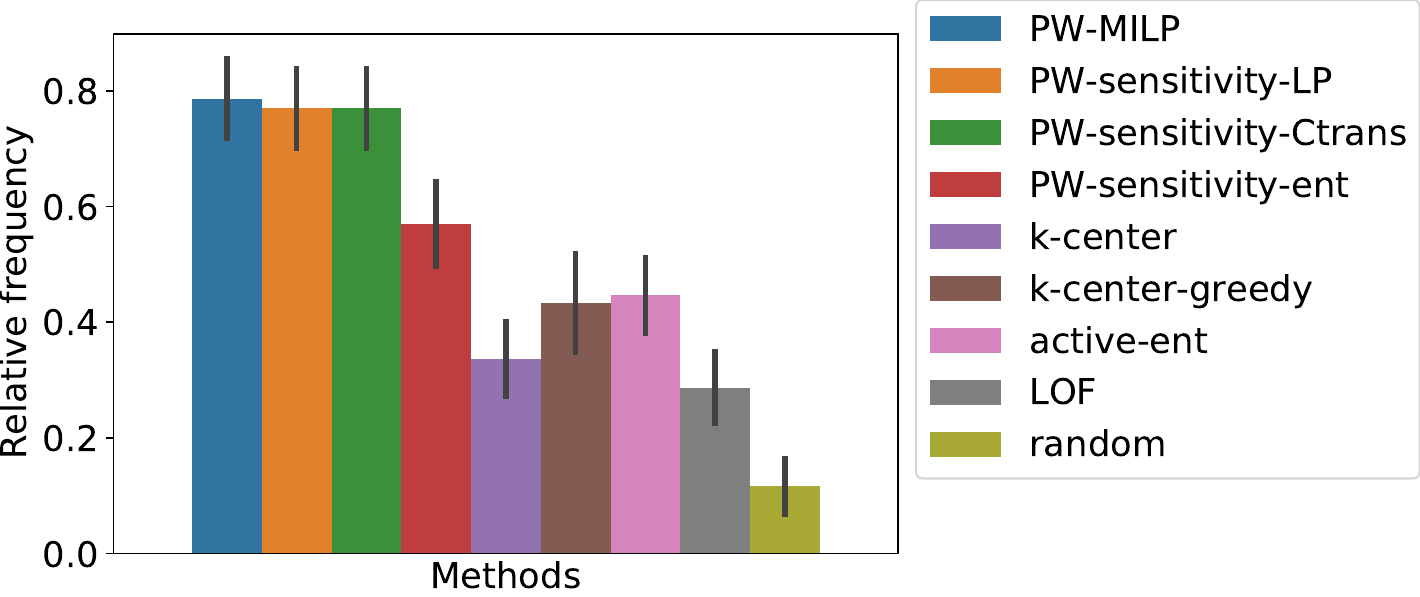}
    \caption{Relative frequency of completely missing category in selected data (MNIST dataset). The black bars denote the standard deviations.}
    \label{fig:mnist-supp}
\end{figure}

In Sec.~\ref{sec:mnist}, we consider the situation, where the frequency of a label in the development dataset is very rare compared to that in the application dataset.
We herein consider another situation, where a label is completely missing in the development dataset.
Fig.~\ref{fig:mnist-supp} illustrates the relative frequency of missing category in selected data.
In this situation, partial Wasserstein covering methods (i.e., \texttt{PW}-$\star$) outperform the baselines including the active learning methods (i.e., $k$-center, $k$-center-greedy, and active-ent) and the anomaly detection method (i.e., LOF).
% We also show the relative frequency for the case where a small amount of data with missing category is included in the development data (i.e., the same setting in the main paper) in Fig.~\ref{fig:mnist-supp}(b) 

\subsection{Finding a missing category on CIFAR-10}

\begin{figure}[t]
    \centering
    \includegraphics[width=0.5\linewidth]{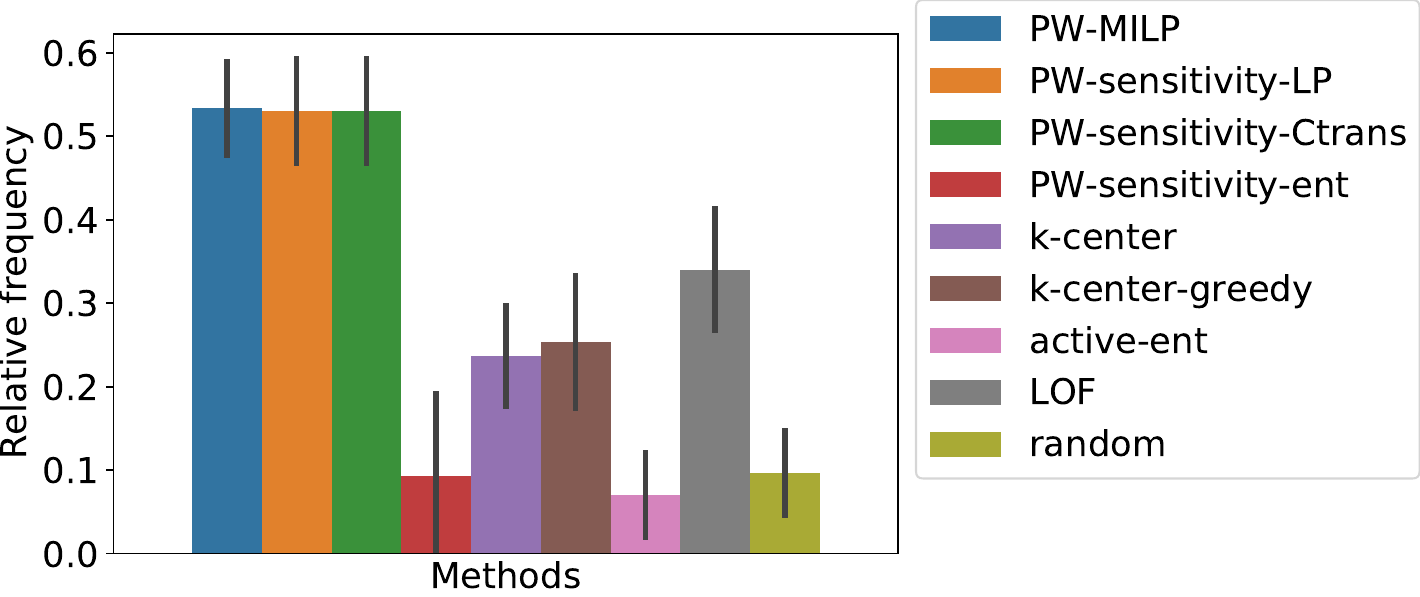}
    \caption{Relative frequency of missing category in selected data (CIFAR-10 dataset). The black bars denote the standard deviations.}
    \label{fig:cifar}
\end{figure}

We perform the missing category finding experiment using CIFAR-10 dataset.
We herein consider \emph{airplane} label (the first label in the CIFAR-10) as the missing label.
As the ground metrics, we employ $L^2$-norm between the feature vectors extracted by a neural network.
We obtain the neural network from torchhub\footnote{\url{https://github.com/chenyaofo/pytorch-cifar-models}}, and train using the subset of the training data of CIFAR-10 from scratch using the Adam optimizer\footnote{We employed the default parameters in PyTorch for the Adam optimizer.} during 10 epochs.
For the subset of the training data, we remove the missing category data so that the percentage of the missing category would be the same as the development data (i.e., 0.5\%).
The other settings are the same as in Sec.5.2.
Fig.~\ref{fig:cifar} illustrates the results indicating that our covering algorithms outperform the baseline, while \texttt{PW-sensitivity-ent} failed to find the missing category. This could be due to the Sinkhorn iterations not sufficiently converged.

\subsection{Missing scene extraction in real driving datasets varying number of data}

In the Sec.~\ref{sec:driving}, we showed the results of 1,500 data for the development dataset, and 3,000 data for the application dataset. We herein show some additional results varying number of data.
We also show top-8 images for each.
These result show that partial Wasserstein covering successfully extract the major differences (i.e., night scenes).

\begin{figure}[t]
    \centering
    \subfigure[Partial Wassersein covering (PW-sensitivity-LP)]{
        \includegraphics[width=\linewidth]{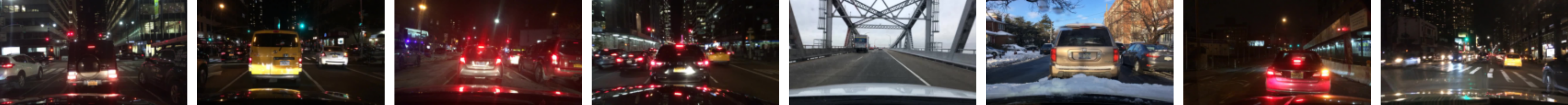}
    }
    \\
    \subfigure[LOF]{
        \includegraphics[width=\linewidth]{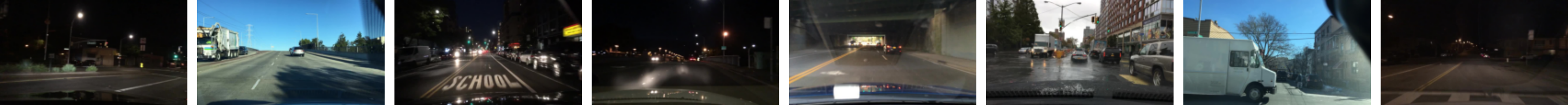}
    } \\
    \subfigure[Coreset (k-center greedy)]{
        \includegraphics[width=\linewidth]{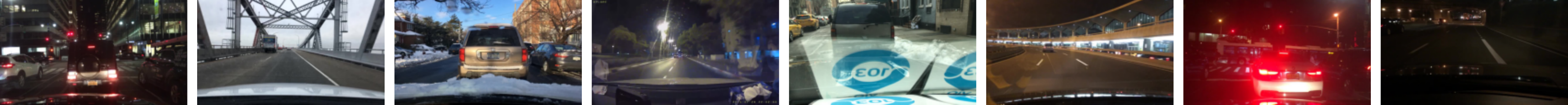}
    }\\
    \subfigure[Random]{
        \includegraphics[width=\linewidth]{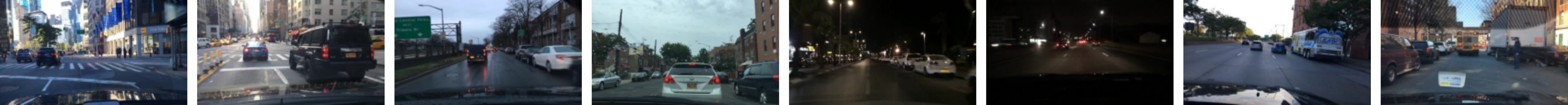}
    }
    \caption{Development dataset 500 and application dataset 1,000}
    \label{fig:bdd-supp1}
\end{figure}

\begin{figure}[t]
    \centering
    \subfigure[Partial Wassersein covering (PW-sensitivity-LP)]{
        \includegraphics[width=\linewidth]{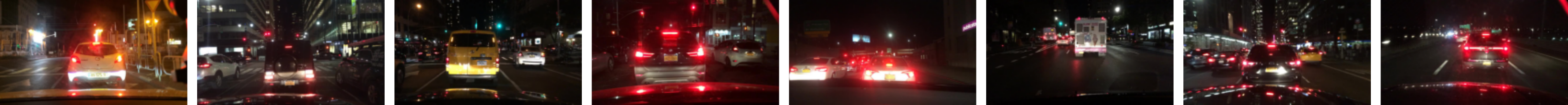}
    }
    \\
    \subfigure[LOF]{
        \includegraphics[width=\linewidth]{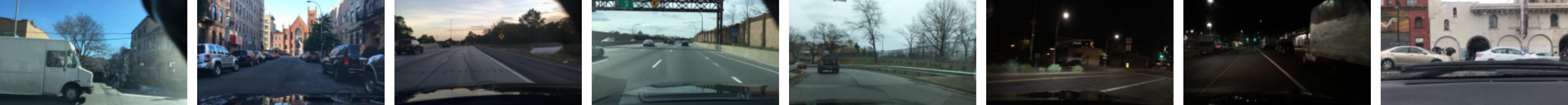}
    } \\
    \subfigure[Coreset (k-center greedy)]{
        \includegraphics[width=\linewidth]{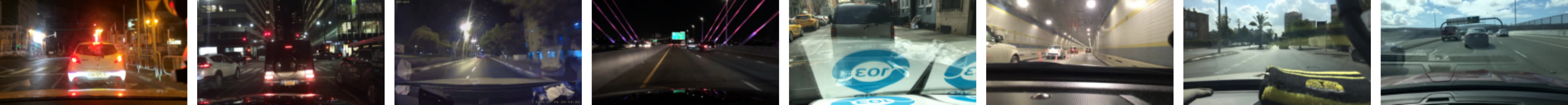}
    }\\
    \subfigure[Random]{
        \includegraphics[width=\linewidth]{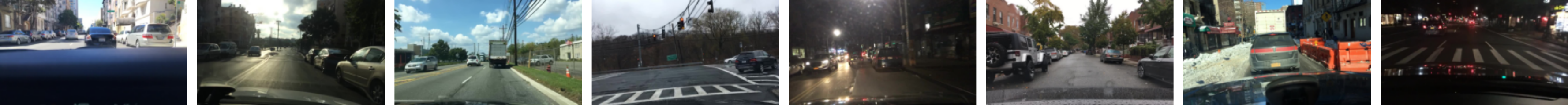}
    }
    \caption{Development dataset 1,000 application dataset 2,000}
    \label{fig:bdd-supp2}
\end{figure}

\begin{figure}[t]
    \centering
    \subfigure[Partial Wassersein covering (PW-sensitivity-LP)]{
        \includegraphics[width=\linewidth]{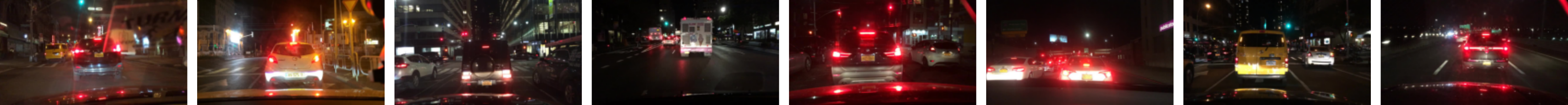}
    }
    \\
    \subfigure[LOF]{
        \includegraphics[width=\linewidth]{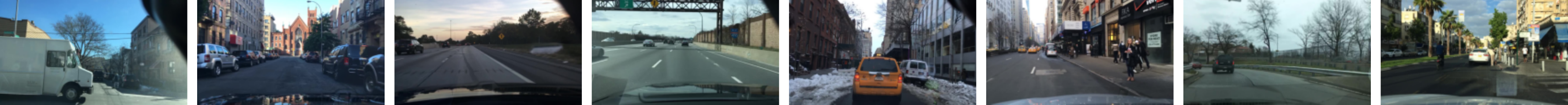}
    } \\
    \subfigure[Coreset (k-center greedy)]{
        \includegraphics[width=\linewidth]{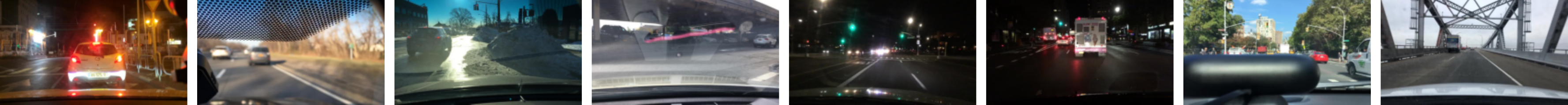}
    }\\
    \subfigure[Random]{
        \includegraphics[width=\linewidth]{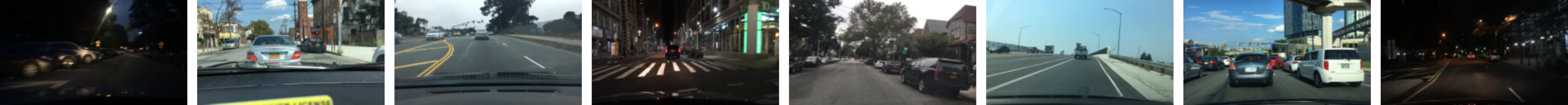}
    }
    \caption{Development dataset 1,500 application dataset 3,000}
    \label{fig:bdd-supp3}
\end{figure}

\section{Discussion}
\label{sec:discussion}

\subsection{General partial OT and unbalanced OT}
We only consider the formulation of the partial Wasserstein divergence (Eq.~1), while the other formulations of the optimal transport can be considered.
We note that the general partial OT~\cite{figalli2010optimal} for discrete distributions is actually the same as Eq.(1) when $\mathbf a^\top \mathbbm 1_n \leq \mathbf b^\top \mathbbm 1_m$.
A covering problem using the unbalanced OT (i.e.,  $\min_{\mathbf P} \left< \mathbf P,  C\right> + \tau_1 D_1(\mathbf P \mathbbm 1_m|\mathbf a) + \tau_2 D_2(\mathbf P^\top \mathbbm 1_n | \mathbf b)$) can be considered, while the submodularity may not be hold in this case.
We also emphasize that we need to adjust the strength of the penalty term (i.e., $\tau_1, \tau_2$) to select data points within the budgets.

\end{document}